\documentclass[final,3p,times,twocolumn]{elsarticle}

\usepackage{amssymb}
\usepackage{amsmath}
\usepackage{booktabs}
\usepackage{multirow}
\usepackage{xurl}
\usepackage{hyperref}

\journal{Knowledge-Based Systems}

\begin{document}

\begin{frontmatter}


\title{Privacy-Preserving RAG by Concealing Sensitive \\ Information from External LLMs}

\author[1]{Saleh Almohaimeed}
\author[2]{Saad Almohaimeed}
\author[1]{Mousa Jari}
\author[1]{Fahad Alotaibi}
\author[1]{Khalid A. Alobaid}
\author[1]{Mohamad Mahmoud Al Rahhal}


%%\author[label1,label2]{}

%%\author{nnn} %% Author name

\affiliation[1]{
    organization={College of Applied Computer Science, King Saud University},
    city={Riyadh},
    country={Saudi Arabia}
}

\affiliation[2]{
    organization={Department of Digital Transformation, Institution of Public Administration},
    city={Riyadh},
    country={Saudi Arabia}
}

%% Abstract
\begin{abstract}
Retrieval-Augmented Generation (RAG) is widely used to improve the performance of Large Language Models (LLMs) in answering user queries. Existing privacy research on RAG has focused on preventing unauthorized users from accessing sensitive data. However, another important problem that is often overlooked in RAG privacy research is that external generators have access to the query and the retrieved documents, which may contain confidential information that could potentially be misused or accessed for unintended purposes. In this paper, we introduce the Sensitive Entity Alias Generator (SEAG), a privacy-preserving framework that empowers users to utilize powerful third-party generators without disclosing sensitive information. SEAG introduces a lightweight model that locates sensitive entities, generates corresponding aliases, and constructs an entity replacement table. The table is used to replace sensitive words in the user's query and in the retrieved documents before they are forwarded to an external generator. For this purpose, two datasets were constructed: one for fine-tuning SEAG models to generate entity replacement tables, and another for evaluating the entire SEAG framework. The experimental results demonstrate the success of the SEAG framework. As for the User metric, which measures the ability of the model to provide a correct response to the user while hiding sensitive information from the external generator, all SEAG models achieved over 80\% accuracy. Additional analysis further evaluated the ability of SEAG models Qwen-3, LLaMA-3.2, and Phi-4 to hide all sensitive entities within given documents. The results show good performance with total accuracies of 77.83\%, 76.73\%, and 74.91\%, respectively.

\end{abstract}

%% Keywords
\begin{keyword}
Retrieval-Augmented Generation (RAG); Privacy-Preserving AI; Large Language Models (LLMs); Sensitive Entity Replacement; Parameter-Efficient Fine-Tuning (PEFT)



\end{keyword}

\end{frontmatter}



\section{Introduction}
\label{sec1}
Retrieval-augmented generation (RAG) has been shown to improve the performance of Large Language Models (LLMs). Specifically, RAG enables LLMs to respond to user queries more effectively by combining LLM's internal knowledge with external knowledge derived from external data sources. The two main components of a RAG system are the retriever, which is responsible for retrieving relevant information, and the generator, which uses the information retrieved to generate accurate responses. This makes LLM a highly effective tool for company search, customer support, and question answering systems.

In spite of this, RAG experiences many privacy issues, which can be seen from two perspectives. First, user threat, as users may have access to sensitive information from external data sources for which they are not authorized. This issue has been widely discussed in many research papers \cite{1st_intro}\cite{2nd_intro}\cite{3rd_intro}. Specifically, the user may deliberately or unintentionally submit questions that lead the RAG system to retrieve and reveal sensitive information from external data sources. Consequently, RAG generates responses that contain confidential data such as financial details \cite{4th_intro}, patient medical records \cite{5th_intro}, or contract information \cite{6th_intro} that the user is not authorized to access.

The second privacy issue that has been overlooked by previous studies is the trustworthiness of the LLM generators. In some applications, the user may have full access to all data in external documents, such as when a physician queries their own patient records. However, the external third-party generator (such as GPT-5 \cite{GPT-5} or Claude-4 sonnet \cite{Claude-4}) may still access and misuse the information. Several articles have reported this risk \cite{c_intro}, \cite{b_intro}, \cite{a_intro}. This threat can be mitigated by installing the generator locally, which ensures that sensitive data remain within the organization and are not exposed to third-party providers.

However, setting up a high-performance LLM locally can be very costly. For example, the cost of an NVIDIA H100 \cite{7th_intro} GPU begins at approximately \$25,000. The deployment of an LLM with more than 100 billion parameters typically requires several H100 GPUs. Furthermore, GPU costs are only one component of the overall cost of deployment. There are also other considerations, such as inference latency, number of concurrent users, networking equipment, and ongoing maintenance. This prevents many organizations from hosting high-performance LLM locally. Alternatively, small LLMs with parameters less than 30 billion can be deployed. However, the efficiency of the generator will be significantly reduced.

In this paper, we propose the Sensitive Entity Alias Generator (SEAG) framework, which enables organizations to leverage powerful external third-party large language models (LLMs), such as GPT-5 \cite{GPT-5} and Claude-4 sonnet \cite{Claude-4}, while deploying only a lightweight local model with approximately 3–4 billion parameters.
As illustrated in Figure \ref{fig1}, a traditional Retrieval-Augmented Generation (RAG) pipeline works as follows: when a user submits a query, the retriever identifies and returns the most relevant documents from external data sources. The query and the retrieved documents are then sent directly to the generator, which produces the final response.

In contrast, our SEAG framework introduces an additional component between the retriever and the generator. After retrieval, both the user query and the relevant documents are passed through the SEAG model, which creates an entity replacement table containing aliases for sensitive entities. Using this table, all sensitive entities in the query and retrieved documents are appropriately replaced with their corresponding aliases.

The anonymized query and documents are then forwarded to the external third-party generator. Since the SEAG model is fine-tuned to preserve semantic meaning and maintain consistency across the documents, the generator can process the information without being aware that sensitive entities have been replaced.

Once the generator produces a response, the output is passed through the entity replacement table once again. Any aliases appearing in the generated response are mapped back to their original entities, resulting in a final answer that preserves both the utility of the external LLM and the privacy of sensitive information.

Many methods can be used to hide sensitive words within a document. Nevertheless, in our work, we fine-tune a model to locate sensitive entities and generate a new alias for each entity while maintaining consistency. In order to accomplish this, we generate two datasets. One for fine-tuning the LLMs, and the other to evaluate the entire framework. The first dataset consists of 640 samples from 8 domains, and the second consists of 600 samples from 6 domains. The evaluation domains are different from those used during fine-tuning, except for two (Legal and Finance). With this setup, we can compare the performance of the fine-tuned models on domains it was trained on against domains not encountered during fine-tuning.

Three LLMs (Qwen-3 \cite{Qwen-3}, LLaMA-3.2 \cite{LLaMA-3.2}, and Phi-4 \cite{phi-4}) have been finetuned using a parameter-efficient fine-tuning (PEFT) technique, called QLoRA \cite{QLora}, all of which are of size 3B with the exception of Qwen-3 \cite{Qwen-3}, which is of 4B. The framework is then evaluated on two popular models, the GPT-5 \cite{GPT-5} and Claude-4 sonnet \cite{Claude-4}. Our SEAG framework is measured by two main metrics. The first metric is the Privacy metric, which measures whether the model correctly hides sensitive information by replacing them with meaningful words, and the results indicated that the best model was LLaMA-3.2 as a finetuned model, which scored 89.67\%. The second metric is named User metric, which measures whether the whole SEAG framework has been implemented successfully. This means the sensitive information has been hidden and the correct results have been shown to the user. Based on this metric, the best results were achieved by the combination of LLaMA-3.2 as the SEAG model and Claude-4-sonnet as the generator, with a score of 83.67\%.

Furthermore, we conducted a study to determine the performance of each fine-tuned SEAG model on hiding sensitive information, concluding that Qwen-3 \cite{Qwen-3} and LLaMA-3.2 \cite{LLaMA-3.2} perform comparably, scoring 77.83\% and 76.73\%, respectively. Lastly, there are two major limitations that are demonstrated in the limitation section, including the length of the documents and the number of sensitive entities in the document.

Our main contributions are:
\begin{itemize}
\item	To the best of our knowledge, we were the first to introduce a framework that protects private documents from being shared with the external third-party generator within RAG systems.

\item	Construct two datasets: The first for fine-tuning SEAG models in order to locate sensitive entities within given documents and to generate semantically consistent aliases. The second is for evaluating the SEAG framework under three different metrics.

\item	Fine-tune three models, Qwen-3, LLaMA-3.2, and Phi-4 on our first dataset, and compare their performance in hiding sensitive information.

\item	Evaluation of the SEAG framework using two third-party generators GPT-5 and Claude-4 sonnet.

\end{itemize}

\section{Related Work}
\label{sec2}

In this section, we will present previous studies related to our work from three perspectives. First, the RAG application and its privacy issues. Second, the previous solutions done to identify sensitive entities within the text and the proposed solution. Third, the parameter-efficient fine-tuning of LLMs.

\subsection{RAG application and its privacy issues}
In recent years, RAG has been widely deployed in many organizations and private entities, as it can be applied to any application with some stored data, such as finance \cite{4th_intro}, medical \cite{5th_intro}, and legal applications \cite{6th_intro}. 

There are many privacy issues associated with the RAG framework; the most common issue is the leakage of sensitive private data. In \cite{1st_intro}, \cite{5th_RW}, and \cite{6th-RW}, they propose different attack methods to extract sensitive information from databases. In \cite{1st_intro} they craft a structured, well-designed prompt for attacking patients’ medical records. In the HealthCareMagic \cite{7th_RW} dataset, using 250 prompts, they were able to extract 89 medical dialogue chunks. In \cite{5th_RW}, they use an agent-based attack that generates adversarial prompts iteratively for extracting data from private RAG knowledge bases. In \cite{6th-RW}, they design a prompt-injected extraction attack, which tells the LLM to reproduce retrieved documents, enabling large-scale leakage of data. Furthermore, there are other attacks that do not use a structured prompt, such as membership inference attack \cite{8th_RW}\cite{9th_RW}\cite{6th-RW}, these attacks do not directly extract documents. Instead, they infer whether a specific record or document is present in the RAG knowledge base. Another attack is the Data Poisoning Attacks \cite{11_RW}\cite{12_RW}, these are knowledge corruption attacks on RAG, where an attacker could inject a few malicious texts into the knowledge database of a RAG system to induce an LLM to generate an attacker-chosen target answer for an attacker-chosen target question.

\subsection{Sensitive Entity Identification}
On the other hand, there are many efforts made to fight against these attacks and protect sensitive information in the retrieved documents within the RAG systems. A popular technique is using synthetic data \cite{1st_intro}\cite{13_RW}\cite{14_RW}. In \cite{13_RW}, they generate a privacy-preserving version of the original data and only provide the synthetic version to the LLM. They use a two-stage generation and refinement paradigm called SAGE, in stage-1 they utilize an attribute-based extraction and generation approach to generate the synthetic data. Then, they propose an agent-based iterative refinement approach to enhance the protection of private information. Their results demonstrate that using their generated synthetic data as RAG data achieves comparable performance to using the original data while effectively mitigating the associated privacy issues. Another solution is the application of the Differential Privacy technique \cite{16_RW}\cite{17_RW}, which protects sensitive information in the document by giving the LLM generator only a privacy-protected list of important concepts from the retrieved document. One major issue that has not been adequately addressed in previous work is the handling of sensitive documents in RAG systems. For example, a document may contain a patient's medical records, and an authorized user, such as a doctor, needs to query this information. However, we may not want the external generator LLM to directly access the sensitive content. In our work, we designed the SEAG framework to handle these issues while returning high-quality answers. 

\subsection{Parameter-efficient fine-tuning of LLM}
After the tremendous adoption of RAG in many applications around the world, many researchers have explored improving RAG efficiency using Parameter-Efficient Fine-Tuning (PEFT) techniques, which improve domain adaptation while minimizing computational costs. Traditional fine-tuning approaches require updating all model parameters, which becomes prohibitively expensive for large language models (LLMs). PEFT methods, such as LoRA \cite{Lora} and QLoRA \cite{QLora}, address this limitation by updating only a small subset of trainable parameters while keeping the original model weights frozen. The most common case is where they fine-tune the generator LLM in the RAG component in their own domain \cite{18_RW}\cite{19_RW}\cite{20_RW}, as an example in \cite{18_RW}, where they fine-tuned 5 different models on a healthcare dataset. The results reveal that all fine-tuned models outperform traditional RAG. 

Another research direction focuses on fine-tuning the retriever component instead of the generator. The overall performance of a RAG system heavily depends on the quality of retrieved documents. Improving the retriever can directly enhance answer accuracy. As a previous work in the OpenRAG paper \cite{21_RW}, they fine-tune the retriever based on the downstream generation objective rather than relying solely on traditional retrieval relevance metrics. In our paper, we took a different approach where we did not focus on the retriever or the generator. To enhance privacy in RAG systems and achieve our objective, we fine-tune an LLM to act as a sensitive information entity recognition and replacement. The retriever will return the document, then it will pass through our fine-tuned LLM to hide sensitive information before the document is being used by the generator to get the answer.

\section{Datasets}
\label{sec3}

In the beginning, two datasets were constructed. First, we constructed a dataset to evaluate our main objective: fine-tuning LLMs to hide sensitive information in texts by replacing sensitive entities with alternatives that are semantically relevant. By doing so, we aim to prevent the reader LLM (Generator) from being aware that these entities have been substituted.

The second dataset was created for the purpose of evaluating the entire SEAG framework. First, the question and its associated documents are given to our fine-tuned SEAG model, that produces an entity replacement table. In both the question and the documents, sensitive entities are located and replaced based on the entity replacement table before being passed to the generator (GPT-5 \cite{GPT-5} or Claude-4 sonnet \cite{Claude-4}). Then, when the generator produces an answer, the entity replacement table is used again to return any substituted entities in the generated answer to their original values. Appendix \ref{tab:phi_replacement_table} contains a sample of entity replacement table.

The second dataset is used to assess the proposed SEAG framework from two perspectives. Firstly, the ability of the fine-tuned SEAG models to conceal sensitive information within the given text. Secondly, the ability of the external generator model to produce the correct answer based on the modified question and document.

\subsection{Dataset for LLMs Fine-tuning}
To construct the first dataset, we collected samples of documents from 8 different domains (news, legal, medical, finance, biology, chemistry, history, and resume). In such a way that every domain has 80 samples, resulting in 640 samples. Then, DeepSeek-V3 \cite{DeepSeek-V3} is prompted to find entities within the given text and substitute them with semantically similar alternative words. We chose DeepSeek-V3 because it has demonstrated a good understanding of the Arabic language in comparison with other models in many research papers \cite{ai_text_detection_ALM} \cite{benchmarking_RAG_AL}.

A set of criteria is used to non-parametrically restrict DeepSeek-V3 \cite{DeepSeek-V3} in the generation of the entity replacement table. There has been a strong emphasis on extracting all name entities, not just a few. Moreover, all replacement words should be consistent. For example, if there are two entities Saudi Arabia and Riyadh, which is the capital of Saudi Arabia. Then when DeepSeek-V3 replaces Saudi Arabia with Qatar, it must also replace Riyadh with the capital city of Qatar, which is Doha. Another criterion: if an entity appears in multiple forms, such as John, John Jordan, and John’s, the entity replacement table must preserve those forms consistently. Create corresponding replacements for each variation, such as replacing John with Mike, John Jordan with Mike Cruz, and John’s with Mike’s. Lastly, the values in the original key must precisely and syntactically match the corresponding references in the document. For example, if a word is mentioned as John’s, then the original key must have the exact word, not jhons, with small letters or John without an apostrophe and letter 's' at the end. 

This set of criteria was not selected arbitrarily; we did this after so many refinements, first asking DeepSeek-V3 \cite{DeepSeek-V3} to generate the entity replacement table, then checking for errors and repeating the process until we reached a stable situation where the generated entity replacement table was sufficient. Most importantly, we did not use a zero-shot prompt for DeepSeek-V3. We included an example question, document, and replacement table in the prompt as we realized that it would enhance the quality of the generated entity replacement table.  

\begin{figure*}[h]
\centering
\includegraphics[width=\textwidth]{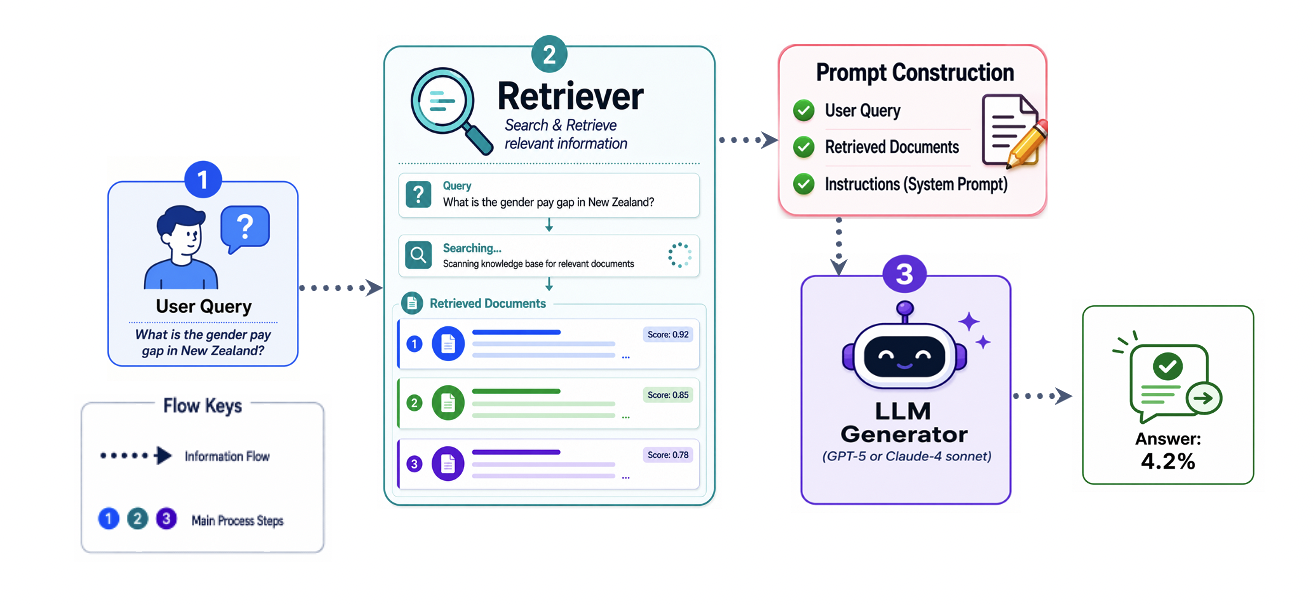}
\caption{The diagram illustrates the structure of a traditional RAG system. It consists of three main components. First, user, who submits a natural language query to the system. Next, the system will use the retriever to extract the most relevant documents to the query from data sources. Following that, the RAG system will forward the query, relevant documents, and predefined system instructions to the generator in order to obtain the appropriate results.}\label{fig1}
\end{figure*}

\subsection{Dataset for evaluating the overall framework}

To evaluate the overall SEAG framework, a second dataset was constructed from six different domains (economic, education, cars, legal, finance, business). Each domain contains 100 samples, resulting in a total of 600 samples. Two members of our team manually constructed the questions for this dataset. This was done to ensure that the question itself contains private entities and also asks about specific results in such a way that the answer should contain private information. The entities include numerical values, dates, names of objects, personal identifiers, etc. Additionally, the number of documents in the second dataset is 300, while the number of questions is 600, which means that there are two questions per document. 

This second dataset was not only built to evaluate the overall SEAG framework, but also to evaluate the performance of the fine-tuned models (SEAG models), to test their ability on hiding sensitive information. We have taken a considerable amount of time to identify each entity within the second dataset documents and evaluate whether the SEAG models were able to hide all of them successfully. The datasets and code are available on {Github.com/Saleh-Almohaimeed/SEAG}.

\section{Methodology}
\label{sec4}

\subsection{Traditional RAG}

Figure \ref{fig1} shows the traditional Retrieval-Augmented Generation (RAG) system, including three main components: (1) User Query, (2) Retriever, and (3) Generator.

(Step 1), The user submits a query in natural language, for instance, "What is the gender pay gap in New Zealand?". The query is then passed to the Retrieval component (Step 2), which uses the existing knowledge base or document repository to retrieve the most relevant documents. The retriever assigns relevance scores to the retrieved documents then selects the top-k documents that are most likely to contain the information required to answer the user's query.

After that, the RAG system constructs a prompt based on three elements: user query, retrieved documents, and a set of system instructions that control the large language model behavior. The prompt provides additional information to the generator, allowing it to base its response on the retrieved information rather than relying completely on its internal knowledge.

In (Step 3) in Figure \ref{fig1}, the constructed prompt is forwarded to the generator LLM, such as GPT-5 \cite{GPT-5} or Claude-4 sonnet \cite{Claude-4}, which integrates the retrieved information to produce the final answer. By incorporating external knowledge during inference, the RAG system mitigates hallucinations, increases factual accuracy, and enables the LLM to answer questions based on information that was not available during pre-training. In the example shown in Figure \ref{fig1}, the model correctly generates the answer "4.2\%" based on the retrieved documents.

\subsection{Sensitive Entity Alias Generator}

\begin{figure*}[t]
\centering
\includegraphics[width=\textwidth]{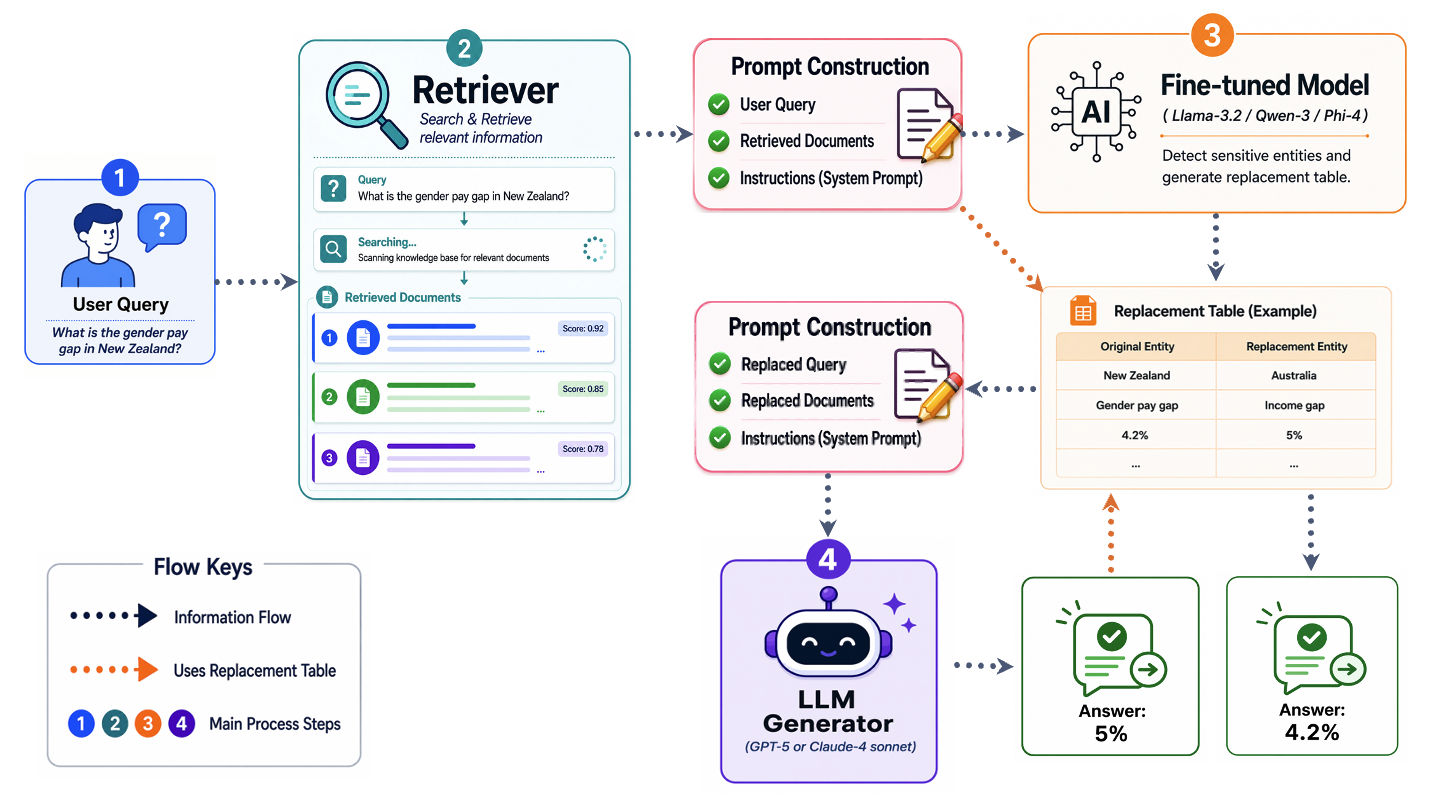}
\caption{The structure of a traditional RAG system alongside our proposed SEAG framework. It consists of four main components. First, the user, who submits a natural language query to the system. Next, the system will use the retriever to extract the most relevant documents to the query from data sources. Following that, the SEAG framework will forward the query and relevant documents to the SEAG model, which will generate an entity replacement table. The table contains all sensitive entities within the query and documents, alongside appropriate aliases for these entities. After that, the SEAG framework will use this table to generate new versions of the query and documents that contain no real sensitive information and forward them to the generator. Finally, the generator will generate an answer, and if the answer contains sensitive entity values, using the entity replacement table, these values will be restored to their original values. Note: orange dotted arrows indicate the stages in which the entity replacement table is used.}\label{fig2}
%% https://en.wikibooks.org/wiki/LaTeX/Importing_Graphics#Importing_external_graphics
\end{figure*}

The proposed SEAG framework is illustrated in Figure \ref{fig2}. As an extension of the traditional RAG pipeline, the framework utilizes a fine-tuned entity replacement model between prompt construction and the external LLM generator. The purpose of this model is to prevent sensitive information in the user query and retrieved documents from being exposed directly to the external generator, while still enabling the system to generate an accurate response. There are four main components in SEAG framework: (1) User Query, (2) Retriever, (3) SEAG model, and (4) Generator.

Initially, the user submits a natural language query (Step 1). As the example shown in Figure \ref{fig2}, the user submits: “What is the gender pay gap in New Zealand?”. The query is then sent to the retriever (Step 2), which will retrieve the most relevant documents from the data source. The documents retrieved are ranked according to their relevance scores and the top-k documents are chosen. (Steps 1 and 2) are the same as in traditional RAG. Step (3) of traditional RAG will take the original query, the retrieved documents, and predefined system instructions and pass them to the generator. However, the user query and retrieved documents may contain sensitive information that should not be shared with an externally hosted LLM (Generator). For instance, they may contain references to New Zealand, gender pay gap, and the value 4.2.

In the SEAG framework, in (Step 3) the original prompt is not sent directly to the generator. Instead, it is sent to a locally deployed fine-tuned model. The fine-tuned model is based on a relatively small language model, such as LLaMA-3.2 \cite{LLaMA-3.2}, Qwen-3 \cite{Qwen-3}, or Phi-4 \cite{phi-4}. It is primarily responsible for identifying sensitive entities and generating appropriate replacements.

The fine-tuned model examines both the user's query and the retrieved documents to identify entities that need to be protected. After that, it creates an entity replacement table that maps every original entity with a corresponding substitute (alias). In the example, New Zealand is replaced with Australia, gender pay gap is replaced with income gap, and 4.2\% is replaced with 5\%. The entity replacement table is constructed in such a manner that the grammatical structure and semantic relationships within the query and documents are maintained.

The entity replacement table is important because it ensures that the same original entity is consistently replaced throughout the query as well as throughout all retrieved documents. As an example, every occurrence of New Zealand must be replaced with Australia, rather than using different country names in different parts of the prompt. In the same way, all related numerical values and other entities should be replaced consistently. In this way, the external generator is able to reason over the transformed information without gaining access to the original sensitive information.

After the entity replacement table has been generated, it is applied to the original prompt. This produces a second, privacy-preserved prompt containing the replaced query, the replaced retrieved documents, and the original system instructions. In Figure \ref{fig2}, orange dotted arrows indicate the stages in which the entity replacement table is used.

In (Step 4), the replaced prompt is sent to an external third-party generator, such as GPT-5 \cite{GPT-5} or Claude-4 sonnet \cite{Claude-4}. The generator receives only the altered version of the information, therefore it does not have direct access to the original sensitive information. The generator then returns an intermediate result. In the illustrated example, it returns 5\%, which is correct with respect to the replaced prompt.

However, this intermediate answer is not returned to the user directly. In the reverse direction, the entity replacement table is used again by the SEAG framework. Each replacement entity appearing in the generated answer is mapped back to its original counterpart. As a result, in Figure \ref{fig2} the intermediate answer of 5\% is restored to its original value of 4.2\%. After the answer has been restored, it will be delivered to the user.

There is one main condition that must be met in order for the SEAG framework to be fully effective. The fine-tuned model must correctly identify and replace all sensitive entities consistently. A missing entity may expose private information, while an incorrect or inconsistent replacement may alter the meaning of the prompt.

\section{Experiments}
\label{sec5}

\subsection{Experimental Setup}

In order to ensure a fair comparison, all three SEAG models were fine-tuned using identical training hyperparameters. We used the AdamW optimizer with a learning rate of $5\times10^{-4}$, a batch size of 8, and trained each model for five epochs. It is true that model-specific hyperparameter tuning can potentially enhance individual performance, however, a fixed configuration was chosen in order to isolate the impact of the internal model architecture on the performance of SEAG model. A single NVIDIA T4 GPU was used for all experiments, and the Hugging Face transformers and PEFT libraries were used to implement the experiments.

\subsection{Evaluation Metrics}
Three different metrics have been used to evaluate the SEAG framework. Firstly, the \textbf{ORG} metric measures the ability of the generator (GPT-5 \cite{GPT-5} or Claude-4 sonnet \cite{Claude-4}) to produce the correct answer when given the original question and documents without using entity aliases. Our goal with this metric is to demonstrate that the questions and documents are not ambiguous and can be answered easily, which helps us identify whether our framework performs well. However, if the questions were difficult to answer, we were not able to determine from the other metrics (\textbf{Privacy} and \textbf{User}) whether the performance decrease was due to the difficulty of the questions or the inefficiency of the framework.

The second metric is \textbf{Privacy}, which measures whether the entities required for answering the question have been successfully hidden. As an example, if the ground-truth answer John cannot be answered by the generator LLM, then this is considered a successful hiding of the entity John. The third metric is \textbf{User}, which evaluates whether the correct answer is displayed to the user when the SEAG framework has been applied. It means that the ground truth answer is presented to the end user after it has been hidden from the external generator. For example, if the ground-truth answer is John, the generator answers with John, and the answer presented to the user is also John it is not considered correct. However, if the ground-truth answer is John, the generator answers with another name like Mike, and the answer presented to the user is John, then the answer is considered correct. In such a way John has been successfully replaced with another alias (Mike) and then restored back to John.

\subsection{Overall Results}

\begin{table*}[t]
\centering
\caption{Performance of different Sensitive Entity Alias Generator (SEAG) models with Claude-4 sonnet and GPT-5 as the generator. \textbf{ORG} measures the ability of the generator to produce the correct answer when given the original question and documents without any changes. \textbf{Privacy} measures whether the entities necessary for answering the question have been successfully hidden. \textbf{User} assesses whether the SEAG framework has been applied successfully and that the correct result is shown to the user while the generator sees the aliases instead. Note: The results of ORG metric are not related to the SEAG models, only to the generators.}
\label{tab:privacy_results}
\renewcommand{\arraystretch}{1.3}

\begin{tabular*}{\textwidth}{@{\extracolsep{\fill}}llccc@{}}
\toprule
\textbf{Generator} & \textbf{SEAG Model} & \textbf{ORG} & \textbf{Privacy} & \textbf{User} \\
\midrule

\multirow{3}{*}{Claude-4 sonnet}
& Qwen-3    & 99.33 & 88.83 & 83.00 \\
& LLaMA-3.2 & 99.33 & 89.67 & 83.67 \\
& Phi-4     & 99.33 & 87.17 & 81.17 \\

\midrule

\multirow{3}{*}{GPT-5}
& Qwen-3     & 99.67 & 87.50 & 83.00 \\
& LLaMA-3.2  & 99.67 & 88.50 & 83.00 \\
& Phi-4     & 99.67 & 84.83 & 80.17 \\

\bottomrule
\end{tabular*}
\end{table*}

The results in Table~\ref{tab:privacy_results} show that the generators GPT-5 and Claude-4 sonnet achieved very high ORG scores. ORG scores ranged from 99.33\% to 99.67\%, which indicates that the original questions and documents are clear and easy for the generators to answer without entity replacement. This confirms that the evaluation dataset questions are not difficult, and that any differences in performance with respect to Privacy and User metrics can be attributed more to the SEAG framework than to the difficulty of the questions.

According to the results of the Privacy metric, all three SEAG models were able to hide sensitive entities successfully in most cases. With Claude-4 sonnet as the generator, LLaMA-3.2 achieved the highest Privacy score of 89.67\%, followed by Qwen-3 with 88.83\%, and Phi-4 with 87.17\%. Similar trends were observed with GPT-5, where LLaMA-3.2 achieved the highest Privacy score (88.50\%), followed by Qwen-3 (87.50\%) and Phi-4 (84.83\%). The results indicate that LLaMA-3.2 produces the most effective entity replacements, making it more difficult for external third-party generators to recover the original sensitive entities.

In the User metric, an evaluation is made of whether the correct final answer is returned to the user following the restoration of the hidden entities. With Claude-4 sonnet, LLaMA-3.2 achieved the highest User score (83.67\%), followed closely by Qwen-3 (83.00\%), while Phi-4 obtained 81.17\%. In GPT-5, both Qwen-3 and LLaMA-3.2 achieved 83.00\%, whereas Phi-4 achieved 80.17\%. Based on these results, it can be concluded that the SEAG framework is capable of maintaining quality of answers for the end user, while protecting sensitive entities from external third-party generator.

Overall, LLaMA-3.2 consistently provided the highest Privacy scores while maintaining high User accuracy across both generators. Furthermore, similar results obtained with Claude-4 sonnet and GPT-5 demonstrate the effectiveness of the SEAG framework across a variety of state-of-the-art generators. Since the Privacy metric evaluates only those entities necessary to answer the user's question, we evaluated all three SEAG models on their capability to hide all named entities in the retrieved documents. The results of this more comprehensive evaluation are presented in the following section.

\subsection{Hiding Sensitive Information}

\begin{table}[t]
\centering
\caption{Entity replacement accuracy across different domains for the evaluated SEAG models. The \textbf{Total Accuracy} is computed over all six domains (600 samples in total), while each domain score is calculated over the entities belonging to that domain.}
\label{tab:domain_results}
\renewcommand{\arraystretch}{1.2}

\begin{tabular}{lccc}
\toprule
\textbf{Domain} & \textbf{Qwen-3} & \textbf{Phi-4} & \textbf{LLaMA-3.2} \\
\midrule
Total Accuracy & 77.83 & 74.91 & 76.73 \\
\midrule
Economic  & 86.01 & 86.11 & 87.46 \\
Education & 75.45 & 71.01 & 74.72 \\
Cars      & 75.14 & 74.94 & 76.16 \\
Legal     & 76.33 & 73.76 & 75.27 \\
Finance   & 77.49 & 73.10 & 74.13 \\
Business      & 77.70 & 72.13 & 74.69 \\
\bottomrule
\end{tabular}
\end{table}

Table~\ref{tab:domain_results} illustrates the entity replacement accuracy across six domains for the SEAG models. Overall, Qwen-3 had the highest overall accuracy (77.83\%), followed by LLaMA-3.2 (76.73\%) and Phi-4 (74.91\%). Since each of the six domains contains 100 samples with approximately 15 sensitive entities per sample, the reported domain accuracy reflects the proportion of correctly replaced sensitive entities in that domain. Interestingly, despite the fact that the SEAG models were fine-tuned using financial and legal data, neither of these domains achieved the highest performance. In contrast, economic consistently produced the best results for all three models, with accuracy ranging from 86.01\% to 87.46\%.

When we have manually examined the dataset, it appears that the economic domain is dominated by factual entities whose boundaries are explicit and whose semantic categories are clearly identifiable. Examples include countries (New Zealand, United States), international organizations (OECD, IMF), dates and years (April 2025), percentages (10\%, 15\%), monetary amounts (USD 900 billion), and well-defined programs such as NextGenerationEU. As these entities usually appear as single nouns, phrases or numerical expressions with little ambiguity, SEAG models can accurately identify and replace them while preserving their surrounding context.

However, the legal and finance domains contain entities that are more context-specific and structurally complex. Legal documents frequently include long legal or policy titles, and hierarchical references. (e.g., Medium-Term Fiscal-Structural Plan 2025--2029 (MTFSP), and National Recovery and Resilience Plan (NRRP)), where accurately determining the complete entity span is more challenging than replacing a standalone country or year.

Similarly, finance documents contain institution-specific organizations, financial reports, and monetary expressions that are often used together within the same statement, such as McKinsey, PWC, S\&amp;P Global, CFA Institute. Thus, in order to perform a correct replacement, it is necessary to simultaneously recognize multiple related entities while maintaining their semantic relationships. Therefore, the superior performance on the economic domain is likely due to its use of standardized, explicitly outlined entities rather than any advantages resulting from domain-specific fine-tuning.

\section{Limitations}

\subsection{Document Length and Entity Density}

Our work results are based on documents with an average length of approximately 300 words, each containing approximately 15 sensitive entities. This size is representative of many practical RAG applications, in which the retrieved chunks are often concise to fit within the model context while still providing sufficient information to answer user questions. However, the proposed framework has not been evaluated on documents that are substantially longer and contain a large number of entities. As document length increases, identifying, replacing, and consistently restoring all sensitive entities may become more difficult, potentially affecting both privacy preservation and the quality of answers. For this reason, future work should explore the scalability of the SEAG framework on larger documents with multiple sensitive entities in order to gain a better understanding of its robustness in more challenging situations.

\label{sec6}

\section{Conclusion}
\label{sec7}

In this paper, we present the Sensitive Entity Alias Generator (SEAG), a privacy-preserving framework for RAG systems. Unlike previous approaches, in which the primary objective is to prevent unauthorized users from accessing sensitive information, the SEAG framework focuses on preventing external third-party generators from accessing sensitive information. In order to accomplish this, the SEAG model is locally deployed between the retriever and the generator, where it detects sensitive entities within the query and documents. Then, generate semantically consistent aliases for these entities. This will ensure that the external third-party generator will only see the aliases and not the original sensitive entities.

To evaluate the proposed framework we constructed two datasets: one to fine-tune a lightweight LLM for finding sensitive entities in given documents and the other to evaluate the entire SEAG framework across multiple domains.

Experimental results show that the proposed framework successfully hides sensitive information while maintaining high answer quality. In comparison to other models, LLaMA-3.2 achieved the best overall performance, achieving the highest Privacy score and maintaining a high User accuracy across both GPT-5 and Claude-4 sonnet generators. 

Also, another experiment was conducted to evaluate each SEAG model across all entities in the retrieved documents. Results indicate that LLaMA-3.2 and Qwen-3 achieve very close performance, with a slight advantage being given to Qwen-3. However, Phi-4 performance was clearly less than the others as it achieved 2\% lower results across all six domains.

Despite the promising results of the proposed framework, several challenges remain. Future research should examine longer documents, documents containing substantially larger numbers of sensitive entities, additional categories of sensitive information, multilingual settings, as well as stronger replacement strategies capable of handling more complex contextual dependencies. Taking these actions will improve the scalability and robustness of RAG systems.

\bibliographystyle{elsarticle-num}
\bibliography{ref}

@inproceedings{1st_intro,
  title     = {The Good and the Bad: Exploring Privacy Issues in Retrieval-Augmented Generation (RAG)},
  author    = {Zeng, Shenglai and Zhang, Jiankun and He, Pengfei and Xing, Yue and Liu, Yiding and Xu, Han and Ren, Jie and Wang, Shuaiqiang and Yin, Dawei and Chang, Yi and Tang, Jiliang},
  booktitle = {Findings of the Association for Computational Linguistics: ACL 2024},
  pages     = {4505--4524},
  address   = {Bangkok, Thailand},
  publisher = {Association for Computational Linguistics},
  year      = {2024},
  doi       = {10.18653/v1/2024.findings-acl.267}
}

@article{2nd_intro,
  title   = {Trustworthiness in Retrieval-Augmented Generation Systems: A Survey},
  author  = {Zhou, Yujia and Zhang, Wenbo and Shao, Jingying and Liu, Yan and Li, Xiaoxi and Jin, Jiajie and Qian, Hongjin and Liu, Zheng and Li, Chaozhuo and Zhang, Jason Chen and Dou, Zhicheng and Yu, Philip S. and Mao, Jiaxin},
  journal = {arXiv preprint arXiv:2409.10102},
  year    = {2024},
  doi     = {10.48550/arXiv.2409.10102}
}

@article{3rd_intro,
  title   = {Privacy Protection in RAG: A Novel Method and Evaluation Framework},
  author  = {Zhang, Yuan and Wu, Jionghan and Li, Rui and Zhang, Tong and Song, Yujie and Li, Chuanyi and Wang, Shangqi and Shen, Hao and Yin, Jiao and Ge, Jidong and Luo, Bin},
  journal = {Information Processing \& Management},
  volume  = {63},
  pages   = {104505},
  year    = {2026},
  doi     = {10.1016/j.ipm.2025.104505}
}

@article{4th_intro,
  title   = {{FairRAG}: A Privacy-Preserving Framework for Fair Financial Decision-Making},
  author  = {Nagpal, Rashmi and Usua, Unyimeabasi and Palacios, Rafael and Gupta, Amar},
  journal = {Applied Sciences},
  volume  = {15},
  number  = {15},
  pages   = {8282},
  year    = {2025},
  doi     = {10.3390/app15158282}
}

@inproceedings{5th_intro,
  title     = {Privacy-Preserving Medical Advising System on Mobile Devices: On-Device PHI Anonymization, Medical Report Retrieval, and Cloud-Based RAG},
  author    = {Weerasekara, Thejan Bandara and Chandeepa, Chamika and Amarasuriya, Oshada S. and Hettiarachchi, Chathura},
  booktitle = {Proceedings of the ACM/IEEE International Conference on Connected Health: Applications, Systems and Engineering Technologies},
  pages     = {447--452},
  publisher = {Association for Computing Machinery},
  year      = {2025},
  doi       = {10.1145/3721201.3725431}
}

@article{6th_intro,
  title   = {Enhancing the Precision and Interpretability of Retrieval-Augmented Generation (RAG) in Legal Technology: A Survey},
  author  = {Hindi, Mahd and Mohammed, Linda and Maaz, Ommama and Alwarafy, Abdulmalik},
  journal = {IEEE Access},
  volume  = {13},
  pages   = {46171--46189},
  year    = {2025},
  doi     = {10.1109/ACCESS.2025.3550145}
}

@misc{c_intro,
  author       = {Itoi, Nikki Goth},
  title        = {Be Careful What You Tell Your {AI} Chatbot},
  year         = {2025},
  month        = oct,
  howpublished = {Stanford Institute for Human-Centered Artificial Intelligence},
  url          = {https://hai.stanford.edu/news/be-careful-what-you-tell-your-ai-chatbot},
  note         = {Accessed: 12 August 2026}
}

@misc{b_intro,
  author       = {{Financial Times}},
  title        = {Florida Sues {OpenAI} and Sam Altman for ``Hurting'' Children},
  year         = {2026},
  month        = jun,
  howpublished = {Financial Times},
  url          = {https://www.ft.com/content/7998db0a-22d1-4b80-b861-2e8b2448e97f},
  note         = {Accessed: 12 August 2026}
}

@misc{a_intro,
  author       = {Pollina, Elvira and Armellini, Alvise},
  title        = {Italy Fines {OpenAI} over {ChatGPT} Privacy Rules Breach},
  year         = {2024},
  month        = {dec},
  howpublished = {Reuters},
  url          = {https://www.reuters.com/technology/italy-fines-openai-15-million-euros-over-privacy-rules-breach-2024-12-20/},
  note         = {Accessed: 12 August 2026}
}

@misc{7th_intro,
  author       = {{NVIDIA Corporation}},
  title        = {{NVIDIA H100 GPU}},
  howpublished = {NVIDIA},
  url          = {https://www.nvidia.com/en-us/data-center/h100/},
  note         = {Accessed: 12 August 2026}
}

@article{5th_RW,
  title   = {Feedback-Guided Extraction of Knowledge Base from Retrieval-Augmented {LLM} Applications},
  author  = {Jiang, Changyue and Pan, Xudong and Hong, Geng and Bao, Chenfu and Chen, Yang and Yang, Min},
  journal = {arXiv preprint arXiv:2411.14110},
  year    = {2024},
  doi     = {10.48550/arXiv.2411.14110}
}

@inproceedings{6th-RW,
  title     = {Follow My Instruction and Spill the Beans: Scalable Data Extraction from Retrieval-Augmented Generation Systems},
  author    = {Qi, Zhenting and Zhang, Hanlin and Xing, Eric P. and Kakade, Sham M. and Lakkaraju, Himabindu},
  booktitle = {The Thirteenth International Conference on Learning Representations},
  year      = {2025},
  publisher = {International Conference on Learning Representations},
  url       = {https://openreview.net/forum?id=Y4aWwRh25b}
}

@misc{7th_RW,
  author       = {Li, Yunxiang and Li, Zihan and Zhang, Kai and Dan, Ruilong and Jiang, Steve and Zhang, You},
  title        = {{ChatDoctor}: A Medical Chat Model Fine-Tuned on a Large Language Model Meta-AI ({LLaMA}) Using Medical Domain Knowledge},
  year         = {2023},
  howpublished = {GitHub repository},
  publisher    = {GitHub},
  url          = {https://github.com/Kent0n-Li/ChatDoctor},
  note         = {Accessed: 12 August 2026}
}

@article{8th_RW,
  title     = {{RAG}-leaks: Difficulty-Calibrated Membership Inference Attacks on Retrieval-Augmented Generation},
  author    = {Wang, Guangshuo and He, Jiajun and Li, Hao and Zhang, Min and Feng, Dengguo},
  journal   = {Science China Information Sciences},
  volume    = {68},
  number    = {6},
  pages     = {160102},
  year      = {2025},
  month     = may,
  publisher = {Springer Nature},
  doi       = {10.1007/s11432-024-4441-4}
}

@inproceedings{9th_RW,
  title     = {Mask-Based Membership Inference Attacks for Retrieval-Augmented Generation},
  author    = {Liu, Mingrui and Zhang, Sixiao and Long, Cheng},
  booktitle = {Proceedings of the ACM Web Conference 2025},
  series    = {WWW '25},
  pages     = {2894--2907},
  year      = {2025},
  month     = apr,
  address   = {Sydney, NSW, Australia},
  publisher = {Association for Computing Machinery},
  location  = {New York, NY, USA},
  isbn      = {979-8-4007-1274-6},
  doi       = {10.1145/3696410.3714771}
}

@inproceedings{11_RW,
  title     = {{PoisonedRAG}: Knowledge Corruption Attacks to {Retrieval-Augmented} Generation of Large Language Models},
  author    = {Zou, Wei and Geng, Runpeng and Wang, Binghui and Jia, Jinyuan},
  booktitle = {34th USENIX Security Symposium (USENIX Security 25)},
  pages     = {3827--3844},
  year      = {2025},
  month     = aug,
  address   = {Seattle, WA},
  publisher = {USENIX Association},
  isbn      = {978-1-939133-52-6},
  url       = {https://www.usenix.org/conference/usenixsecurity25/presentation/zou-poisonedrag}
}

@article{12_RW,
  title     = {{Phantom}: General Backdoor Attacks on Retrieval Augmented Language Generation},
  author    = {Chaudhari, Harsh and Severi, Giorgio and Abascal, John and Suri, Anshuman and Jagielski, Matthew and Choquette-Choo, Christopher A. and Nasr, Milad and Nita-Rotaru, Cristina and Oprea, Alina},
  journal   = {ACM Transactions on AI Security and Privacy},
  year      = {2026},
  month     = mar,
  publisher = {Association for Computing Machinery},
  doi       = {10.1145/3796729},
  url       = {https://doi.org/10.1145/3796729}
}

@inproceedings{13_RW,
  title     = {Mitigating the Privacy Issues in Retrieval-Augmented Generation ({RAG}) via Pure Synthetic Data},
  author    = {Zeng, Shenglai and Zhang, Jiankun and He, Pengfei and Ren, Jie and Zheng, Tianqi and Lu, Hanqing and Xu, Han and Liu, Hui and Xing, Yue and Tang, Jiliang},
  booktitle = {Proceedings of the 2025 Conference on Empirical Methods in Natural Language Processing},
  pages     = {24538--24569},
  year      = {2025},
  month     = nov,
  address   = {Suzhou, China},
  publisher = {Association for Computational Linguistics},
  isbn      = {979-8-89176-332-6},
  doi       = {10.18653/v1/2025.emnlp-main.1247},
  url       = {https://aclanthology.org/2025.emnlp-main.1247/}
}

@inproceedings{14_RW,
  title     = {Large Language Model as Attributed Training Data Generator: A Tale of Diversity and Bias},
  author    = {Yu, Yue and Zhuang, Yuchen and Zhang, Jieyu and Meng, Yu and Ratner, Alexander J. and Krishna, Ranjay and Shen, Jiaming and Zhang, Chao},
  booktitle = {Advances in Neural Information Processing Systems},
  volume    = {36},
  year      = {2023},
  publisher = {Curran Associates, Inc.},
  doi       = {10.52202/075280-2433},
  url       = {https://proceedings.neurips.cc/paper_files/paper/2023/hash/ae9500c4f5607caf2eff033c67daa9d7-Abstract-Datasets_and_Benchmarks.html}
}

@article{16_RW,
  title   = {Differentially Private Retrieval-Augmented Generation},
  author  = {Tang, Tingting and Flemings, James and Wang, Yongqin and Annavaram, Murali},
  journal = {arXiv preprint arXiv:2602.14374},
  year    = {2026},
  month   = feb,
  doi     = {10.48550/arXiv.2602.14374},
  url     = {https://arxiv.org/abs/2602.14374}
}

@inproceedings{17_RW,
  title     = {{RAG} with Differential Privacy},
  author    = {Grislain, Nicolas},
  booktitle = {2025 IEEE Conference on Artificial Intelligence (CAI)},
  pages     = {847--852},
  year      = {2025},
  month     = may,
  publisher = {IEEE},
  url       = {https://ieeexplore.ieee.org/document/11050672}
}

@article{18_RW,
  title     = {Medical {LLMs}: Fine-Tuning vs. Retrieval-Augmented Generation},
  author    = {Pingua, Bhagyajit and Sahoo, Adyakanta and Kandpal, Meenakshi and Murmu, Deepak and Rautaray, Jyotirmayee and Barik, Rabindra Kumar and Saikia, Manob Jyoti},
  journal   = {Bioengineering},
  volume    = {12},
  number    = {7},
  pages     = {687},
  year      = {2025},
  month     = jun,
  publisher = {MDPI},
  doi       = {10.3390/bioengineering12070687},
  url       = {https://doi.org/10.3390/bioengineering12070687}
}

@article{19_RW,
  title   = {{ALoFTRAG}: Automatic Local Fine Tuning for Retrieval Augmented Generation},
  author  = {Devine, Peter},
  journal = {arXiv preprint arXiv:2501.11929},
  year    = {2025},
  month   = jan,
  doi     = {10.48550/arXiv.2501.11929},
  url     = {https://arxiv.org/abs/2501.11929}
}

@article{20_RW,
  title     = {Enhancing Large Language Models for Specialized Domains: A Two-Stage Framework with Parameter-Sensitive {LoRA} Fine-Tuning and Chain-of-Thought {RAG}},
  author    = {He, Yao and Zhu, Xuanbing and Li, Donghan and Wang, Hongyu},
  journal   = {Electronics},
  volume    = {14},
  number    = {10},
  pages     = {1961},
  year      = {2025},
  publisher = {MDPI},
  doi       = {10.3390/electronics14101961},
  url       = {https://doi.org/10.3390/electronics14101961}
}

@inproceedings{21_RW,
  title     = {Open-{RAG}: Enhanced Retrieval Augmented Reasoning with Open-Source Large Language Models},
  author    = {Islam, Shayekh Bin and Rahman, Md Asib and Hossain, K S M Tozammel and Hoque, Enamul and Joty, Shafiq and Parvez, Md Rizwan},
  booktitle = {Findings of the Association for Computational Linguistics: EMNLP 2024},
  pages     = {14231--14244},
  year      = {2024},
  month     = nov,
  address   = {Miami, Florida, USA},
  publisher = {Association for Computational Linguistics},
  doi       = {10.18653/v1/2024.findings-emnlp.831},
  url       = {https://aclanthology.org/2024.findings-emnlp.831/}
}

@article{Qwen-3,
  title   = {{Qwen3} Technical Report},
  author  = {Yang, An and Li, Anfeng and Yang},
  journal = {arXiv preprint arXiv:2505.09388},
  year    = {2025},
  doi     = {10.48550/arXiv.2505.09388},
  url     = {https://arxiv.org/abs/2505.09388}
}

@misc{LLaMA-3.2,
  author       = {{Meta AI}},
  title        = {{Llama 3.2} Model Card},
  year         = {2024},
  publisher    = {Meta},
  url          = {https://github.com/meta-llama/llama-models/blob/main/models/llama3_2/MODEL_CARD.md},
  note         = {Accessed: 12 August 2026}
}

@article{phi-4,
  title   = {{Phi-4-Mini} Technical Report: Compact yet Powerful Multimodal Language Models via Mixture-of-{LoRAs}},
  author  = {Abouelenin, Abdelrahman and Ashfaq, Atabak and Atkinson, Adam and Awadalla, Hany and Bach, Nguyen and Bao, Jianmin and Benhaim, Alon and Cai, Martin and Chaudhary, Vishrav and Chen, Congcong and others},
  journal = {arXiv preprint arXiv:2503.01743},
  year    = {2025},
  doi     = {10.48550/arXiv.2503.01743},
  url     = {https://arxiv.org/abs/2503.01743}
}

@misc{Claude-4,
  author       = {{Anthropic}},
  title        = {Introducing {Claude 4}},
  year         = {2025},
  month        = may,
  howpublished = {Anthropic},
  url          = {https://www.anthropic.com/news/claude-4},
  note         = {Accessed: 12 August 2026}
}

@misc{GPT-5,
  author       = {{OpenAI}},
  title        = {{GPT-5} System Card},
  year         = {2025},
  month        = aug,
  howpublished = {OpenAI},
  url          = {https://openai.com/index/gpt-5-system-card/},
  note         = {Accessed: 12 August 2026}
}

@article{DeepSeek-V3,
  title   = {{DeepSeek-V3} Technical Report},
  author  = {{DeepSeek-AI}},
  journal = {arXiv preprint arXiv:2412.19437},
  year    = {2024},
  doi     = {10.48550/arXiv.2412.19437},
  url     = {https://arxiv.org/abs/2412.19437}
}

@inproceedings{Lora,
  title     = {{LoRA}: Low-Rank Adaptation of Large Language Models},
  author    = {Hu, Edward J. and Shen, Yelong and Wallis, Phillip and Allen-Zhu, Zeyuan and Li, Yuanzhi and Wang, Shean and Wang, Lu and Chen, Weizhu},
  booktitle = {International Conference on Learning Representations},
  year      = {2022},
  url       = {https://openreview.net/forum?id=nZeVKeeFYf9}
}

@inproceedings{QLora,
  title     = {{QLoRA}: Efficient Finetuning of Quantized {LLMs}},
  author    = {Dettmers, Tim and Pagnoni, Artidoro and Holtzman, Ari and Zettlemoyer, Luke},
  booktitle = {Advances in Neural Information Processing Systems},
  volume    = {36},
  pages     = {10088--10115},
  year      = {2023},
  publisher = {Curran Associates, Inc.},
  doi       = {10.52202/075280-0441},
  url       = {https://proceedings.neurips.cc/paper_files/paper/2023/hash/1feb87871436031bdc0f2beaa62a049b-Abstract-Conference.html}
}

@article{ai_text_detection_ALM,
  title     = {{AI} Text Detectors and the Misclassification of Slightly Polished Arabic Text},
  author    = {Almohaimeed, Saleh and Almohaimeed, Saad and Jari, Mousa and Alobaid, Khalid A. and Alotaibi, Fahad},
  journal   = {Journal of Big Data},
  year      = {2026},
  month     = jun,
  publisher = {Springer Nature},
  doi       = {10.1186/s40537-026-01492-8},
  url       = {https://doi.org/10.1186/s40537-026-01492-8}
}

@article{benchmarking_RAG_AL,
  title     = {Benchmarking Retrieval Augmented Generation {LLMs} for Arabic Noise Robustness},
  author    = {Almohaimeed, Saleh and Alabduljabbar, Abdulrahman and Jari, Mousa and Alkhowaiter, Mohammed and Almohaimeed, Saad and Al Rahhal, Mohamad Mahmoud},
  journal   = {Frontiers in Big Data},
  volume    = {9},
  year      = {2026},
  publisher = {Frontiers Media SA},
  doi       = {10.3389/fdata.2026.1884673},
  url       = {https://doi.org/10.3389/fdata.2026.1884673}
}

\appendix
\section{Example of Replacement Table}
\label{app1}

\begin{table*}[t]
\centering
\caption{Example of a replacement table generated by the fine-tuned Phi-4 model.}
\label{tab:phi_replacement_table}
\renewcommand{\arraystretch}{1.15}
\begin{tabular}{lll}
\toprule
\textbf{Entity Type} & \textbf{Original Entity} & \textbf{Replacement Entity} \\
\midrule
Date         & April           & March \\
Location     & New Zealand     & Australia \\
Date         & 2025            & 2026 \\
Organization & OECD            & EU \\
Figure       & Figure 1.1      & Figure 2.2 \\
Location     & Middle East     & South Asia \\
Location     & United States   & Canada \\
Percentage   & 10\%            & 8\% \\
Percentage   & 15\%            & 12\% \\
Date         & August          & July \\
Date         & April 2026   & March 2027 \\
Date         & 2026            & 2027 \\
Commodity    & dairy           & grain \\
Commodity    & beef            & meat \\
Sector       & Tourism         & Manufacturing \\
\bottomrule
\end{tabular}
\end{table*}

 Table \ref{tab:phi_replacement_table} shows an example of an entity replacement table generated by Phi-4 model. The table has the detected sensitive entities, their corresponding entity types, and the generated aliases used to anonymize the user query and retrieved documents before they are sent to the external LLM.
 
\end{document}